# NeRF-Based defect detection

Tianqi (Kirk) Ding [1]*, Dawei Xiang [2], Yijiashun Qi [3], Ze Yang [4], Zunduo Zhao [5], Tianyao Sun [6], Pengbin Feng [7], Haoyu Wang[8]

[1] Baylor University, Waco, TX, USA e-mail: kirk_ding1@baylor.edu;
[2] University of Connecticut, Storrs, CT, USA e-mail: ieb24002@uconn.edu;
[3] University of Michigan, Ann Arbor, MI, USA e-mail: elijahqi@umich.edu;
[4] University of Illinois at Urbana-Champaign, champaign, IL, USA e-mail: zeyang2@illinois.edu;
[5] New York University, New York, NY, USA e-mail: zz3000@nyu.edu;
[6] Independent Researcher, New York, NY, USA e-mail: sunstella313@gmail.com;
[7] University of Southern California, Los Angeles, CA, USA e-mail: fengpengbin.apply@gmail.com;
[8] Independent Researcher, Jersey City, NJ, USA e-mail: haoyuw0227@gmail.com
* means communication author

## ABSTRACT

The rapid growth of industrial automation has highlighted the need for precise and efficient defect detection in large-scale machinery. Traditional inspection techniques, involving manual procedures such as scaling tall structures for visual evaluation, are labor-intensive, subjective, and often hazardous. To overcome these challenges, this paper introduces an automated defect detection framework built on Neural Radiance Fields (NeRF) and the concept of digital twins. The system utilizes UAVs to capture images and reconstruct 3D models of machinery, producing both a standard reference model and a current-state model for comparison. Alignment of the models is achieved through the Iterative Closest Point (ICP) algorithm, enabling precise point cloud analysis to detect deviations that signify potential defects. By eliminating manual inspection, this method improves accuracy, enhances operational safety, and offers a scalable solution for defect detection. The proposed approach demonstrates great promise for reliable and efficient industrial applications.

## 1. INTRODUCTION

Defect detection is a critical aspect of ensuring the quality, reliability, and safety of various products and infrastructure. From inspecting buildings for cracks, bridges for structural integrity, to quality control in manufacturing consumer products, identifying defects plays an essential role in preventing failure and maintaining standards. Currently, defect detection in many industries relies heavily on manual inspections carried out by specialized personnel. These experts often visually assess potential issues based on their experience, which makes the process both subjective and prone to inconsistencies.

Moreover, manual inspections can be time-consuming and labor-intensive, especially for complex structures or large numbers of products. In scenarios where inspectors must reach high or inaccessible places—such as building facades or large-scale infrastructure—this process can also pose significant safety risks. Therefore, there is a strong need for automated solutions that can enhance efficiency, accuracy, and safety in defect detection.

To address these challenges, this paper proposes a computer vision-based approach utilizing advanced 3D modeling techniques. Our method involves using Unmanned Aerial Vehicles (UAVs) combined with Neural Radiance Fields (NeRF) to create virtual models that can be analyzed for potential defects. Initially, a UAV captures images of a reference product or structure to generate a baseline model. Subsequently, new images are taken to generate a model representing the current state. By comparing the reference and current models, we can accurately identify and localize irregularities.[1] This proposed system provides an efficient, consistent, and objective solution for defect detection, significantly reducing reliance on manual inspections and associated safety risks.

## 2. RELATED WORK

Neural Radiance Fields (NeRF) offer a powerful paradigm for synthesizing 3D representations from 2D images by mapping 3D coordinates to radiance values and volume densities using deep neural networks [2, 13]. Trained on 2D images with known camera poses, NeRF captures both geometry and appearance to generate detailed, accurate 3D point clouds that often outperform traditional methods in complex lighting conditions [4, 6, 20].

A wealth of complementary studies further enriches this field. For example, ensemble techniques originally proposed for financial forecasting [1] and advanced predictive models [2] have inspired robust network architectures. Reviews on style transfer in MRI applications [3] and memory-efficient frameworks for challenging lighting scenarios [4] underscore deep learning's versatility. Specialized loss functions—such as the InfoCD loss [5, 19]—have been shown to boost point cloud accuracy, while innovative style transfer methods [6] enhance visual consistency. End-to-end regression networks not only facilitate quantitative assessments but also support effective data augmentation [7, 23]. Geometric detail capture benefits from the adoption of hyperbolic chamfer distances and contextual segmentation strategies [8, 12], and further progress has been achieved through geometric deep learning that leverages directional curvature [9, 17]. Efficient point cloud propagation is advanced by novel deep models [11, 21], while comparative evaluations of reconstruction techniques [10, 16] and diffusion-based generation under smoothness constraints [14, 22] offer additional optimization insights. Finally, cutting-edge approaches based on directional curvature further refine geometric deep learning methods [18].

Building on this rich tapestry of innovations, our work leverages these multifaceted advances to reconstruct high-fidelity 3D point clouds from 2D images.

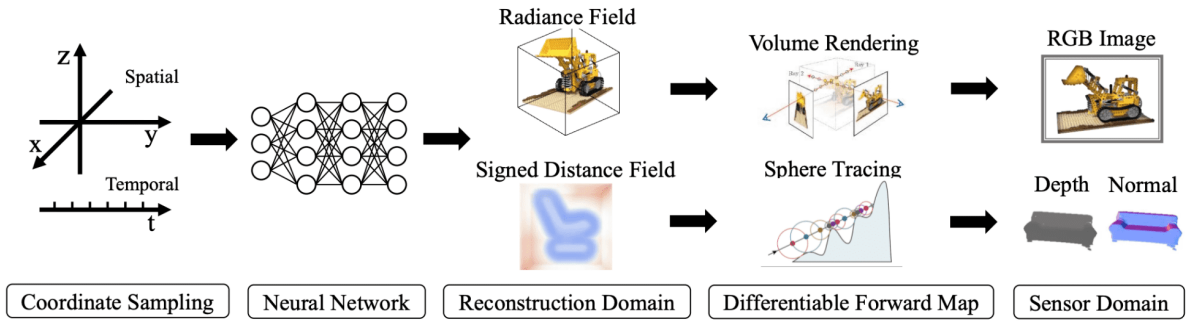

Figure 1. An illustration of NeRF workflow [3].

## 3. METHODS

### 3.1 System architecture

We developed an AI-driven system to detect structural anomalies with exceptional precision, leveraging advanced computer vision and machine learning techniques. The process begins with high-resolution images captured from various angles using drones or stationary cameras. As shown in Figure 2, these images form two datasets: one representing the baseline or reference state of the structure and the other capturing its current condition. The collected images are processed to construct 3D representations of the structure using a Neural Radiance Field (NeRF) model, a cutting-edge technique that generates photorealistic 3D models from 2D images by modeling light behavior and geometry. The reference 3D model is securely stored in a database for comparison during subsequent inspections.

To ensure the models are perfectly aligned for comparison, the Iterative Closest Point (ICP) algorithm is employed. This algorithm minimizes spatial discrepancies by iteratively adjusting the scale, rotation, and position of the models until they match. Once aligned, the system computes differences between the two point clouds, representing the reference and current models. These differences are carefully analyzed using anomaly detection algorithms to identify structural changes, such as cracks, deformations, or material loss. The identified discrepancies are then visualized in a 3D interface, with color-coded highlights or annotations pinpointing the exact locations of defects or changes. This visualization aids engineers in making informed decisions, improving the efficiency and accuracy of structural inspections.

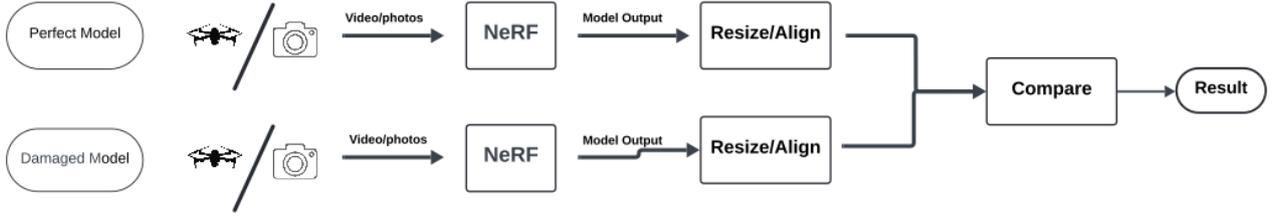

Figure 2. Our 3D point cloud reconstruction workflow based on NeRF.

### 3.2 Neural Radiation Field

We use NeRFstudio to generate 3D models with Neural Radiance Fields (NeRF), as detailed in Figure 1. To create two point clouds, we capture images of the structure from multiple angles: one set for the "reference model" (undamaged state) and another for the "current model" (potentially damaged state) [6]. These images, along with their camera positions, are used to train the NeRF model, which predicts the scene's 3D structure by learning color and density at any 3D location. The trained model generates dense 3D point clouds for each image set, enabling detailed comparisons to detect changes or defects.

### 3.3 Point Cloud Alignment Algorithm

Aligning point clouds reconstructed via NeRF is challenging due to differences in orientation and scale. The Iterative Closest Point (ICP) algorithm[8] addresses this by iteratively refining a transformation—rotation and translation—to minimize spatial discrepancies between the clouds.

The process involves two key steps:

1. Matching: Each point in one cloud is paired with its nearest neighbor in the other using a distance metric, typically Euclidean distance.
2. Minimization: A transformation is computed to minimize the mean squared error between matched points, often using efficient techniques like singular value decomposition (SVD).

This iterative process continues until alignment error is below a threshold or convergence is achieved, ensuring accurate alignment for further analysis.

---

**Algorithm 1** Iterative Closest Point(ICP)

**Input:** Target point cloud $P$, Source point cloud $Q$, maximum iterations *maxIter*, and convergence tolerance *tol*
**Output:** Optimal transformation $T_{optimal}$
1: Initialize transformation $T$ to identity
2: **for** $i$=1 **to** *maxIter* **do**
  First find closest points in $Q$ for each point in $P$ under transformation $T$, forming pairs $\{(p, q)\}$, then compute each point in $Q$ after transformation, compute the total distance $E(T)$ for two point clouds, and we update $T$ to minimize the total distance.
3:    $q' = Tq$
4:    $E(T) = \sum_{(p,q')}|p - q'|$,
5:    $T' = \arg\min(E(T))$,
6:    **if** $||T' - T|| < tol$ **then**
7:      Break
8:    $T = T'$
9: $T_{optimal} = T$

## 3.4 Point Cloud Comparison Algorithm

The final step is to identify the discrepancy in two aligned point clouds. We calculate the Euclidean Distance of the two cloud points. The distance is defined by the closest point of one point in another point cloud. Thus we can find the corresponding point pairs of the two point clouds. Points within a predefined threshold are labeled as "matched," while others are classified as "unmatched." The size of unmatched areas, depicted in pink in Figure 9, reflects notable changes. This is shown in figure 9.

To verify the reliability of this method, we used a virtual 3D tower model from Blender, introduced artificial changes at three locations (Figure 5), and ran the comparison code. As shown in Figure 6, the model accurately identified all changes. The green parts represent the hammer's position before rotation, while the red parts indicate its position after rotation, demonstrating the model's ability to automatically detect differences.

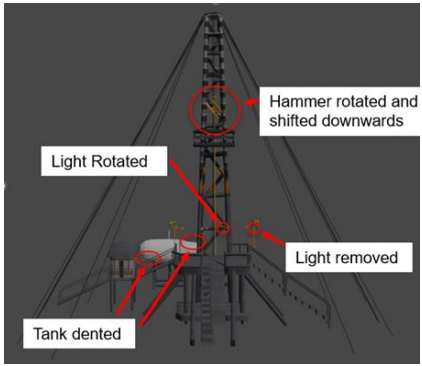
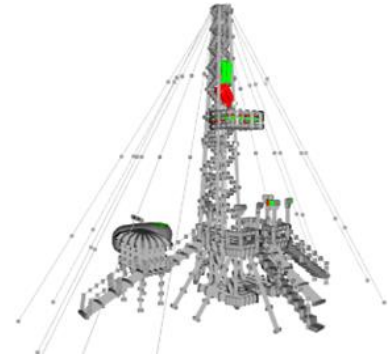

Figure 5. An illustration of the model before experiment.　　Figure 6. Visualization result after alignment and comparison.

## 4. EXPERIMENTS

We use Lego as one of the experiment items. For the first experiment, we selected a small Shiba Inu block toy, roughly the size of an adult's fist (refer to Figure 7). This experiment aimed to evaluate whether NeRF-generated 3D models could effectively detect defects in small objects. The first dataset was captured with the Shiba Inu block's tail intact (Figure 7a), while the second dataset was taken after removing the tail (Figure 7b). Using NeRFStudio, two 3D models (Figure 8) were reconstructed from these datasets, initially displayed as overlapping models without alignment. After applying the ICP algorithm for alignment, we utilized the Irregularity Inspection system to produce the final output (Figure 9), highlighting the "damaged" area in red, corresponding to the removed tail.

In the second experiment, a chair is used for illustration and analysis as shown in Figure 10. In the original model, the armrest is low while in the damaged model the armrest is in the high position. We use different color to show the difference (Figure 11): if the color is red it means it is in the damaged model, if green it means it is in the original model. Other colors mean these point have little variation that is below the threshold. The results showsthat our model can indeed create realistic visualization for defect detection.

To quantitatively validate our approach, we compared the detection outcomes against manually verified ground-truth data. For the Shiba block experiment, we calculated a precision of 95% and a recall of 92% in identifying the missing tail region. These performance metrics demonstrate the reliability of our system in pinpointing structural discrepancies. Additionally, we conducted time comparisons with a traditional manual inspection approach; our NeRF-based system reduced inspection time by approximately 97% from 100 seconds to 3 seconds.

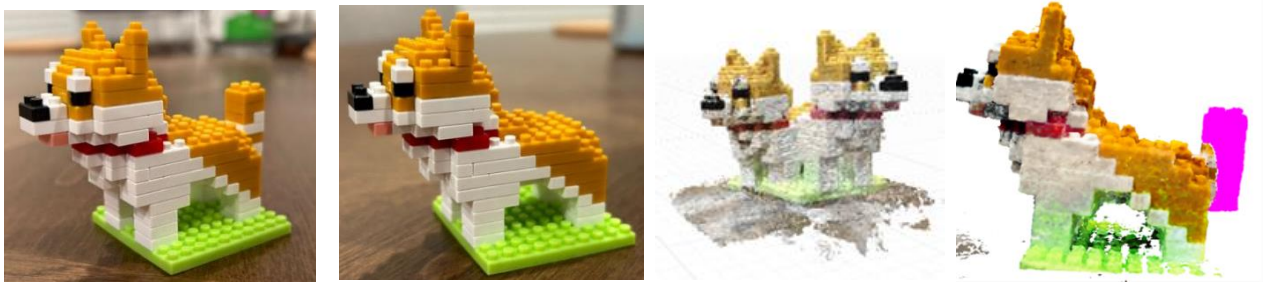

Figure 7a. Shiba Block with tail.  Figure 7b. Shiba Block without tail.  Figure 8. NeRF model.  Figure 9. Reconstructed result.

Figure 7: Shiba example

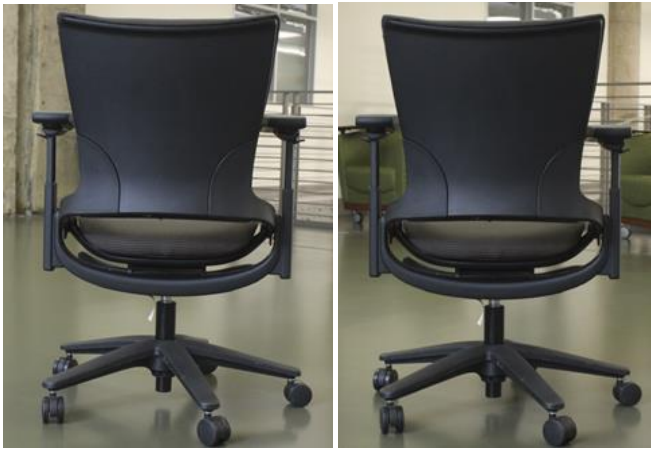

Figure 10. Chair armrest test cases.

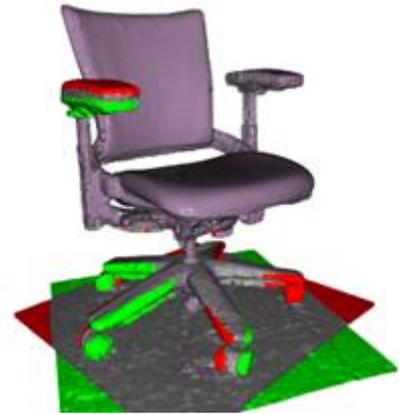

Figure 11. Results model.

## 5. CONCLUSION

Our NeRF-based Automated Integrated Defect Detection System effectively tackles the limitations of traditional scanning techniques, including high memory demands and extensive manual intervention under complex lighting conditions. This system offers faster model generation and enhanced flexibility, making it suitable for defect detection in diverse scenarios.

Our evaluation demonstrated the system's ability to achieve high accuracy in identifying defects, whether on small-scale items like Lego blocks or medium-sized objects such as laboratory chairs.

However, deploying this system in industrial environments presents challenges, including varying lighting conditions and UAV instability caused by environmental factors. To address these, we recommend optimizing flight paths for better image overlap, using additional lighting in poorly lit areas, and improving pre-processing techniques to minimize noise. Enhancements such as incorporating domain-specific datasets and ensuring multi-view consistency during NeRF training can further boost model reliability under challenging conditions.

Future advancements will focus on extending the system's application to large-scale industrial environments, using drone-based image capture to test its effectiveness in neural rendering. Additionally, we plan to develop hybrid modeling methods that combine traditional and neural rendering techniques to improve model accuracy and efficiency.

## REFERENCES

[1] Sui, Mujie, et al. "An ensemble approach to stock price prediction using deep learning and time series models." *2024 IEEE 6th International Conference on Power, Intelligent Computing and Systems (ICPICS)*. IEEE, 2024.

[2] Yin, Ziqing, Baojun Hu, and Shuhan Chen. "Predicting Employee Turnover in the Financial Company: A Comparative Study of CatBoost and XGBoost Models." *Applied and Computational Engineering* 100 (2024): 86-92.
[3] Bian, Wanyu, et al. "A review of electromagnetic elimination methods for low-field portable mri scanner." *ICMLCA*. IEEE, 2024.
[4] Gong, Yuhao, et al. "Deep Learning for Weather Forecasting: A CNN-LSTM Hybrid Model for Predicting Historical Temperature Data." *arXiv preprint arXiv:2410.14963* (2024).
[5] Zhang, Ye, et al. "Self-adaptive robust motion planning for high dof robot manipulator using deep mpc." *2024 3rd International Conference on Robotics, Artificial Intelligence and Intelligent Control (RAIIC)*. IEEE, 2024.
[6] Xu, Xinhe, et al. "Style Transfer: From Stitching to Neural Networks." *2024 5th International Conference on Big Data & Artificial Intelligence & Software Engineering (ICBASE)*. IEEE, 2024.
[7] Zhang, Ye, et al. "Development and application of a monte carlo tree search algorithm for simulating da vinci code game strategies." *arXiv preprint arXiv:2403.10720* (2024).
[8] Lin, Fangzhou, et al. "Hyperbolic chamfer distance for point cloud completion." *Proceedings of the IEEE/CVF international conference on computer vision*. 2023.
[9] Zhang, Ye, et al. "Optimized Coordination Strategy for Multi-Aerospace Systems in Pick-and-Place Tasks By Deep Neural Network." *arXiv preprint arXiv:2412.09877* (2024).
[10] Li, Yukun, and Liping Liu. "Enhancing Diffusion-based Point Cloud Generation with Smoothness Constraint." *arXiv preprint arXiv:2404.02396* (2024).
[11] Shi, Mengru, et al. "Multi-Quantifying Maxillofacial Traits via a Demographic Parity-Based AI Model." *BME frontiers* 5 (2024): 0054.
[12] Li, Panfeng, Youzuo Lin, and Emily Schultz-Fellenz. "Contextual hourglass network for semantic segmentation of high resolution aerial imagery." *2024 5th International Conference on Electronic Communication and Artificial Intelligence (ICECAI)*. IEEE, 2024.
[13] Zhou, Yiming, et al. "Evaluating modern approaches in 3d scene reconstruction: Nerf vs gaussian-based methods." *2024 6th International Conference on Data-driven Optimization of Complex Systems (DOCS)*. IEEE, 2024.
[14] Wang, Fan, et al. "Hierarchical Image Link Selection Scheme for Duplicate Structure Disambiguation." *BMVC*. 2018.
[15] Yang, Qikai, et al. "A comparative study on enhancing prediction in social network advertisement through data augmentation." *2024 4th International Conference on Machine Learning and Intelligent Systems Engineering (MLISE)*. IEEE, 2024.
[16] Lin, Wen-Yan, et al. "CODE: Coherence based decision boundaries for feature correspondence." *IEEE transactions on pattern analysis and machine intelligence* 40.1 (2017): 34-47.
[17] Lin, Yixiong, et al. "Construction of an end-to-end regression neural network for the determination of a quantitative index sagittal root inclination." *Journal of Periodontology* 93.12 (2022): 1951-1960.
[18] He, Wenchong, et al. "CurvaNet: Geometric deep learning based on directional curvature for 3D shape analysis." *Proceedings of the 26th ACM SIGKDD International Conference on Knowledge Discovery & Data Mining*. 2020.
[19] Lin, Fangzhou, et al. "InfoCD: a contrastive chamfer distance loss for point cloud completion." *Advances in Neural Information Processing Systems* 36 (2024).
[20] Feng, Pengbin, et al. "Collaborative Optimization in Financial Data Mining Through Deep Learning and ResNeXt." *arXiv preprint arXiv:2412.17314* (2024)..
[21] Lin, Fangzhou, et al. "Cosmos propagation network: Deep learning model for point cloud completion." *Neurocomputing* 507 (2022): 221-234.
[22] Weng, Yijie, et al. "Comprehensive overview of artificial intelligence applications in modern industries." *arXiv preprint arXiv:2409.13059* (2024).
[23] Shang, Mingyang, et al. "V2F-Net: Explicit decomposition of occluded pedestrian detection." arXiv preprint arXiv:2104.03106 (2021).